%% file: main.tex
\documentclass[10pt,twocolumn,letterpaper]{article}

\usepackage{ragged2e}
\usepackage[pagenumbers]{iccv}
\usepackage[accsupp]{axessibility}

\input{preamble}

\definecolor{iccvblue}{rgb}{0.21,0.49,0.74}
\usepackage[pagebackref,breaklinks,colorlinks,allcolors=iccvblue]{hyperref}
\usepackage{makecell}

\def\paperID{9922}
\def\confName{ICCV}
\def\confYear{2025}
\definecolor{DnCBG}{rgb}{0.9, 0.9, 1.} 

\title{\Model: Non-collapsing Representation Learning with Progressive Freezing}

\author{Goker Erdogan$^{\dag}$ \and Nikhil Parthasarathy$^{\dag}$ \and 
Catalin Ionescu$^{\dag}$ \and Drew A. Hudson$^{\dag}$ \and Alexander Lerchner$^{\dag}$ \and Andrew Zisserman$^{\dag\Diamond}$ \and Mehdi S.\ M.\ Sajjadi$^{\dag}$ \and {Jo\~ao Carreira$^{\dag}$} \\[3mm]
\small{
$^{\dag}$Google DeepMind \;\;\;
$^{\Diamond}$University of Oxford}\\
}

\begin{document}
\maketitle
\input{sec/0_abstract}    
\input{sec/1_intro}
\input{sec/2_related_work}
\input{sec/3_methodology}
\input{sec/4_results}
\input{sec/5_conclusion}
\clearpage
\section*{Acknowledgements}
\begin{FlushLeft}
We would like to thank Zhiwei Deng and Caroline Pantofaru for their helpful comments and feedback on earlier drafts of this paper.
\end{FlushLeft}
{
    \small
    \bibliographystyle{ieeenat_fullname}
    \bibliography{main}
}
\clearpage
\input{sec/6_supplementary}

\end{document}

%% file: preamble.tex
\usepackage{booktabs}
\usepackage{multirow}
\usepackage{pifont}
\usepackage{placeins}
\usepackage{siunitx}
\usepackage{xcolor,colortbl}
\usepackage{algorithm}
\usepackage{algpseudocode}
\usepackage{amsmath}
\usepackage{minted}

\newcommand{\Model}{LayerLock\xspace}
\newcommand{\xmark}{\ding{55}}%

%% file: sec/0_abstract.tex
\begin{abstract}
We introduce \Model, a simple yet effective approach for self-supervised visual representation learning, that gradually transitions throughout training from predicting shallow features to deeper ones through progressive layer freezing. First, we make the observation that during training of video masked-autoencoding (MAE) models, ViT layers converge in the order of their depth: shallower layers converge early, deeper layers converge late. We then show that this observation can be exploited to accelerate standard MAE by progressively freezing the model according to an explicit schedule, throughout training. Furthermore, this same schedule can be used in a simple and scalable approach to latent prediction that does not suffer from ``representation collapse''. We apply our proposed approach, \Model, to both pixel and latent prediction approaches with large models of up to 4B parameters and show improvements on both semantic (action classification) and low level (depth estimation) vision tasks.
\end{abstract}

%% file: sec/1_intro.tex
\section{Introduction}
\label{sec:intro}

The visual world is highly complex and contains regularities across many scales, ranging from low-level spatiotemporal cues about 3D shape and geometry, to highly abstract semantic information like object categories or actions. Biological systems can learn to extract these wide range of regularities in the visual stream in a mostly unsupervised way and hence generalize well to a wide variety of low to high-level visual tasks. Recent advances in self-supervised learning from video \citep{carreira2024scaling, el2024scalable, he2022masked, tong2022videomae, wang2023videomae, bardes2023vjepa, oquab2023dinov2}  bring us closer to building such systems, but still suffer from various issues of data efficiency and training stability among others. In this paper, we present a new self-supervised learning technique that progressively predicts higher-level representations of the input, and leads to a more stable and effective learning that improves performance on a range of low to high-level visual tasks, ranging from depth prediction to action classification.

\begin{figure}[t]
    \centering
    \vspace*{12pt}
    \includegraphics[width=\linewidth]{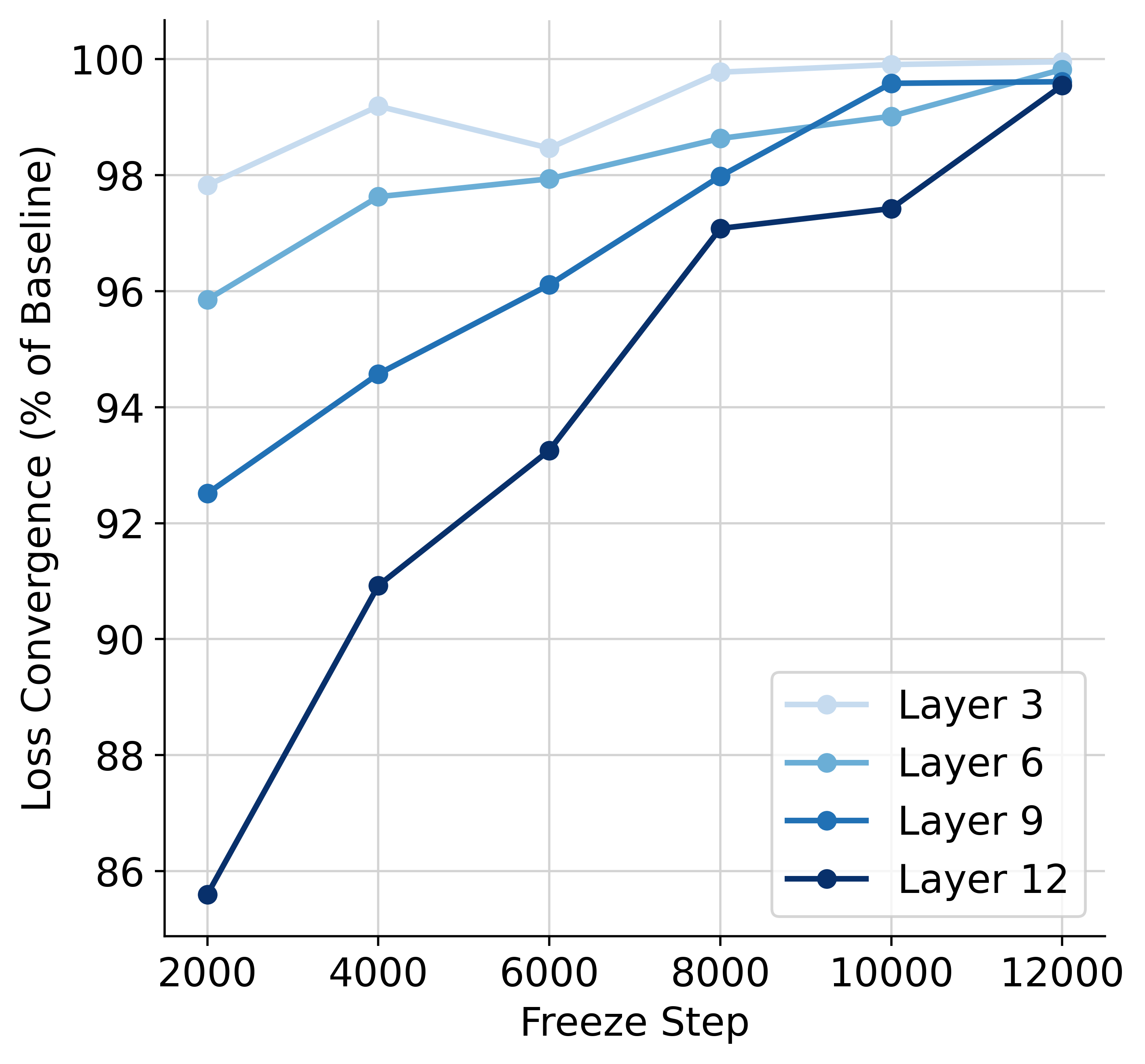}
    \caption{\textbf{In video masked auto-encoding, network layers converge during training in order of their depth.} We measure the final loss ($L_{base}$) of a baseline (unfrozen) model after training for 14000 steps. Each point, similarly shows the corresponding final training loss when freezing the network up to layer $L$ at step $T$. We see that shallow layers ``converge'' faster than deeper layers as they can be frozen at earlier steps while still being able to minimize the final loss very close to $L_{base}$). In other words, layer convergence order is correlated with layer depth, motivating our proposed \textbf{LayerLock} progressive freezing approach to learning.
    }
    \label{fig:convergence_step}
\end{figure}

One way to categorize self-supervised visual representation learning techniques is to look at whether they make predictions in pixel space (i.e., reconstruct the input) or in a learned latent space \citep{lecun2022path, he2022masked, bardes2023vjepa}. Pixel prediction approaches are appealing because reconstruction provides a stable learning signal and grounds the learned representations firmly in the visual stream. However, these approaches tend to require large amounts of training data \citep{wang2023videomae}, and may focus on lower-level visual information that is not always aligned with the downstream tasks we care about \citep{balestriero2024recon, ramesh2024many}. Meanwhile, latent prediction approaches eschew predicting pixels and can learn to capture higher-level visual information and achieve good performance on downstream tasks with much less data \citep{chen2020simple, grill2020byol, oquab2023dinov2, bardes2023vjepa}. However, these approaches often involve various modeling tricks such as asymmetric architectures or target encoders to train without instabilities \citep{wang2022onthe, li2022understanding}.

In this paper, we propose \Model, a new approach that progressively freezes layers and dynamically updates the prediction target to transition from shallow features (pixels or early layer activations) to increasingly deeper intermediate latent model activations. \Model allows us to combine the best of pixel and latent-based approaches, by evolving the prediction target from low-level to high-level visual features throughout training. This progressive freezing approach is partly motivated by the observation that earlier layers in neural networks tend to converge earlier during training (see \cref{fig:convergence_step}). This is akin to critical periods in biological development, where neural structures for specific visual abilities are only plastic for a limited time period during development \citep{levelt2012critical, hubel1959receptive}. By dynamically evolving the prediction target throughout training, we aim to learn better video representations that capture both low and high-level visual information. This further results in a model that is highly stable, does not suffer from representation collapse issues, and allows us to train very large video models (e.g., 4B).

\vspace{1mm}
\noindent \textbf{Contributions.} Our contributions are four-fold: 1) We present \Model, a new simple yet effective technique that progressively freezes layers in a network and predicts a dynamically evolving target from shallow features to deeper and deeper features. 2) We show that this recipe can be applied on both pixel (e.g., VideoMAE, \cite{tong2022videomae, wang2023videomae}) and latent (e.g., V-JEPA \cite{bardes2023vjepa}) prediction approaches for visual representation learning. In both cases, \Model learns video representations that achieve better performance on semantic tasks and low-level dense tasks. 3) We demonstrate that \Model saves memory and compute compared to the vanilla MAE setup, by progressively freezing intermediate layers of the network and hence requiring a backward pass over fewer and fewer layers, as training progresses. 4) We further introduce novel 3D rotary positional embeddings that lead to significant performance improvements across all the explored downstream tasks.

\vspace{1mm}
\noindent \textbf{Paper structure.} We discuss related work in the next section. In \cref{sec:methodology}, we present our approach \Model and discuss training details, baselines, and evaluation tasks. Then, we present our experimental results in Section~\ref{sec:results} and finally conclude in Section~\ref{sec:conclusions}.

%% file: sec/2_related_work.tex
\section{Related work}
\label{sec:related_work}

There is a wide literature on unsupervised learning from images and videos with a wide array of approaches which we roughly categorize into pixel vs. latent prediction.

\vspace{1mm}
\noindent \textbf{Pixel prediction.}
Reconstructing the input image or video, i.e. auto-encoding, has a long history \citep{rumelhart1986learning} and been a popular representation learning technique since the early days of deep learning \citep{kingma2014auto, rezende2014stochastic, masci2011stacked, vincent2010stacked}. While early models were mostly applied to images, there have been various video auto-encoders using convolutional architectures \citep{patraucean2015spatio, lai21video, zhao2017spatio}.  More recently, with the wide adoption of Vision Transformers \citep{dosovitskiy2020image}, masked auto-encoding has emerged as the strongest pixel prediction approach for image and video representation learning \citep{he2022masked, tong2022videomae, wang2023videomae, carreira2024scaling}. However, while pixel prediction generally leads to strong performance, it generally learns slower than latent prediction approaches \citep{wang2023videomae}. There is also work that suggests predicting pixels as an objective is, in fact, not well aligned with all downstream tasks \citep{balestriero2024recon, ramesh2024many}.

\vspace{1mm}
\noindent \textbf{Latent prediction.}
Latent approaches do not predict pixels and instead operate in a learned latent space \citep{balestriero2024recon, lecun2022path}. These approaches are appealing because pixel prediction is in general more difficult and as mentioned above may not align well with downstream tasks. An  important challenge in latent prediction approaches is to avoid \textit{representation collapse}, where the model learns to map all inputs to the same vector or other trivial prediction solution. An influential line of early work pursued contrastive predictive coding \citep{vandenoord2018cpc}; afterwards a wide variety of approaches have been proposed: variants of contrastive learning \citep{chen2020simple, he2020momentum, tian2021dnc}, various regularization techniques to avoid collapse in non-contrastive settings \citep{zbontar2021barlow, bardes2021vicreg, bardes2022vicregl}, clustering-based approaches \citep{caron2018deepcluster, caron2020swav, asano2019sela}, self-distillation  \citep{caron2021emerging, oquab2023dinov2, zhou2021ibot}, and asymmetric non-contrastive approaches \citep{grill2020byol, guo2022byolexplore}. While initially proposed for images, most of these techniques were later applied to videos as well \citep{qian2020cvrl, qian2021spatiotemporal, dave2021tclr, salehi2023timetuning, sermanet2017tcn, ni2022mscl, wang2015unsupervised, parthasarathy2022vito, venkataramanan2023imagenet}.

 We propose here a simple, general, and effective method that gradually progresses from pixel to latent prediction, to achieve the best of both worlds -- grounding on pixels to avoid collapse on the one hand, while progressing towards latents to encourage the emergence of abstract and semantic features on the other.

Note that one component of our approach is the progressive freezing of layers as we switch to deeper prediction targets in the network. A similar progressive freezing approach was investigated by \cite{brock2017freezeout} in an extended abstract for accelerating training while minimizing loss of accuracy in a supervised setting. Note that our approach \Model combines progressive freezing with dynamically changing the prediction targets throughout training and instead focuses more on representation learning in a self-supervised setting rather than accelerating training.

%% file: sec/3_methodology.tex
\section{Methodology}
\label{sec:methodology}
\begin{figure*}[t]
    \centering
    \vspace*{12pt}
    \includegraphics[width=\linewidth]{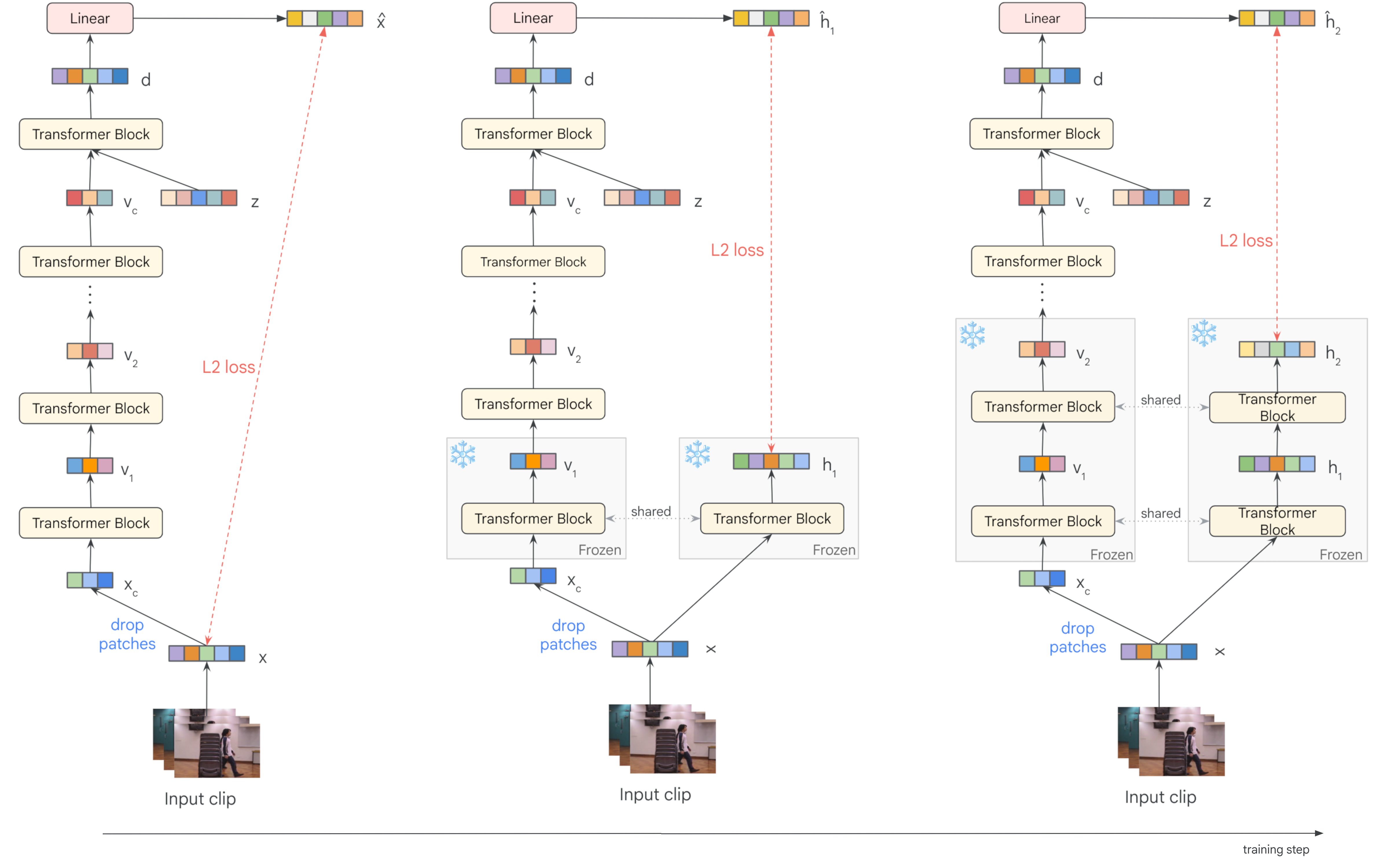}
    \caption{\textbf{Proposed learning paradigm.} Learning goes through multiple stages as the model switches between prediction targets: \textbf{Left:} no frozen layers, predicting pixels $x$, \textbf{Middle:} freezing first layer and predicting output of first layer $h_1$, \textbf{Right:} freezing first two layers and predicting the output of the second layer $h_2$. $z$ are latents added for decoding. For illustration purposes, only a single Transformer block is shown after $v_c$ and $z$ are concatenated. Freezing continues progressively, according to a pre-determined schedule.}
    \label{fig:model}
\end{figure*}

We will explain \Model as is applied on a standard masked-autoencoding (MAE) setup, hence we first outline the vanilla MAE approach, where we feed a masked video clip to the model and aim to reconstruct the input, essentially filling out the masked parts of the video using information from unmasked parts. We then introduce our proposed progressive freezing approach and show how we apply it on the standard MAE setup (see \cref{fig:model} for an illustration of our proposed model and \cref{alg:forward} for pseudocode outlining the forward pass). Finally, in \cref{sec:vjepa_method}, we will explain how we apply \Model to latent prediction approaches like V-JEPA. 

\subsection{Model Architecture}
\label{sec:model}

\vspace{1mm}
\noindent \textbf{Encoding.} Let $x \in \mathbb{R}^{T\times H\times W\times 3}$ be an input video clip of $T$ frames with resolution $H\times W$. We first split this video into patches of size $t \times h \times w$ and flatten it. Then, we drop some of the input patches randomly and linearly project to $D$ dimensions to get our context input $x_c \in \mathbb{R}^{K \times D}$. Note we do not use any form of special masking here such as tube masking and just randomly mask each patch independently of others. After adding positional embeddings to this input, we feed it to a standard Vision Transformer \citep{dosovitskiy2020image, dehghani2023scaling} to get patch embeddings $v_c$.

\vspace{1mm}
\noindent \textbf{Decoding.}
To predict the unmasked input $x$, we first append a grid of decoding latents $z \in \mathbb{R}^{N \times D}$, with one latent for each patch in the input, to the patch embeddings $v_c$. For flexibility and efficiency, instead of using a separate decoder network, we append $z$ before one of the Transformer blocks in the Vision Transformer backbone. By applying the rest of the Transformer blocks over both $v_c$ and $z$, we get the processed latent tokens $d \in \mathbb{R}^{N \times D}$, which are then fed to a patch-wise linear decoder. The patch-wise linear decoder projects each patch from $D$ dimensions to the pixel space, and reshapes the output $\hat{x}$ to the input shape $T\times H\times W\times 3$. The model is trained on a simple L2 loss between unmasked inputs $x$ and predictions $\hat{x}$.

\vspace{1mm}
\noindent \textbf{Positional Embeddings.} For positional embeddings, we use a novel, simple and effective, 3D version of rotary positional embeddings \citep{su2021rope}. We partition the feature dimensions into 4 parts, and apply a separate 1D-RoPE embedding to 3 of these using the positions along the time, height, and width axes of the input video grid respectively, leaving the last part of features without position information. We further find that applying RoPE is less effective in the attention (as initially proposed) than after the first normalization layer at the beginning of the VIT block. We found this novel rotary positional embedding improves performance on downstream tasks (see \cref{sec:results} for more details).

For latent decoding tokens, we use a grid of rotary positional embeddings. Our initial experiments have shown that using learned latent tokens did not improve results, and so we directly use positional embeddings for decoding.

This completes the description of our implementation of a `standard' MAE where the targets are pixels. We next motivate and describe LayerLock.

\subsection{Ordered layerwise convergence in MAEs} 
\label{sec:layer_convergence}
\vspace{1mm}
In order to motivate our proposed methodological changes to the standard MAE paradigm, we first present a key observation regarding the convergence of layers during training. We start by measuring the converged loss $L_{base}$ for the baseline model trained for 14000 steps. We then test the convergence of different layers of a 16 layer MAE based on the following simple principle: a network block up to layer $L$ is considered ``converged'' at a given step $T$ if we can freeze the network up to this layer (at step $T$) and continue training such that the loss is minimized to $L_{base}$. The discrepancy (which we can measure as a percent deviation) between the frozen network loss and $L_{base}$ is then a measure of layer convergence at a certain point in training. 

Shown in \cref{fig:convergence_step}, given this metric, we evaluate different layers $L \in [3, 6, 9, 12]$ and freezing times $T \in [2K, 4K, 6K, 8K, 10K, 12K]$ and find that layer convergence is in fact correlated with network depth: low-level layers converge earlier in training than deeper layers. Based on these observations, we propose that MAE training can be accelerated via a schedule-based progressive freezing approach. The results of this baseline are shown in Sec. \ref{sec:results}. 

\subsection{LayerLock: Progressive freezing with latent prediction}
\label{sec:layerlock_method}
While progressive freezing can acelerate standard MAE training, freezing layers can also act as a way to generate stable latent targets, similar to how EMA networks and stop-gradients are used in methods such as V-JEPA \cite{bardes2023vjepa}. Therefore, our final method, \Model extends simple layer freezing to dynamically transition from predicting shallow features (e.g., pixels or early layer activations) towards deeper latent prediction as layers are frozen. Specifically, let us assume we freeze the first $k$ layers of the model, and predict $h_k$ -- the output of the $k$th layer for the full input $x$. To predict $h_k$, we take the processed latent tokens $d$ and use a new patch-wise linear decoder to project each token to the number of features in the target layer to get predictions $\hat{h_k}$. The model is still trained on a simple L2 loss between targets $h_k$ and predictions $\hat{h_k}$.

For freezing layers, we find the following simple schedule works well. First, we train the model on pixels as targets for $N_{\text{pixel}}$ steps. Then every $N$ steps, we freeze the next $k$ unfrozen layers of the model and predict $h_k$, the output of the $k$th layer for the full input $x$. After $N$ further steps of training, we freeze the next $k$ layers to predict $h_{2k}$, $h_{3k}$. This is repeated, each time using a new learned linear layer to project from tokens to latent space. See \cref{fig:model} for an illustration of this process. Note that for efficiency, one can choose to compute the latent loss only on a subset of the input patches. In our experiments, we show results with both options, i.e., with and without patch dropping for latent loss.

\begin{algorithm*}
\caption{Pseudocode of the forward pass of \Model as applied to MAE. Here \texttt{encoder} refers to the part of the ViT backbone until decoding latents are concatenated for decoding. Similarly, \texttt{decoder} refers to the rest of the backbone. See main text for more details.}
\label{alg:forward}
\begin{minted}[linenos, numbersep=5pt,gobble=2, framesep=2mm, fontsize=\small]{python}
  def forward(video: [B, T, H, W, C], step: int):
    # Get layer to freeze according to schedule. freeze_layer = 0 means no freezing.
    freeze_layer = freeze_layer_schedule(step)
    
    tokens = patchify_and_flatten(video)  # [B, N, D]
    
    # Encode.
    tokens_enc = mask_random(tokens)  # [B, K, D]
    patch_embeddings, _ = encoder(tokens_enc, freeze_layer=freeze_layer)
    
    # Decode.
    decoding_tokens = rope(tokens.shape)  # [B, N, D]
    all_tokens = jnp.concatenate([decoding_tokens, patch_embeddings]
    out_tokens = decoder(all_tokens)[:, 0:N]
    pred_tokens = linear[freeze_layer](out_tokens)  # [B, N, D]
    
    # Compute targets.
    _, layer_outs = encode(tokens)  # Returns list of layer outputs
    layer_outs = [tokens, *layer_outs]  # Prepend pixels as first output
    layer_outs = jax.lax.stop_gradient(layer_outs)
    
    # Calculate loss
    loss = jnp.mean(jnp.square(pred_tokens - layer_outs[freeze_layer])
    return loss
\end{minted}
\end{algorithm*}

\subsection{Training}
\label{sec:training}

First, we demonstrate the effectiveness of \Model on the 4DS model family \cite{carreira2024scaling}, a recent large-scale strong representative model for the video MAE approach \cite{tong2022videomae, wang2023videomae}. In this case, \Model starts by predicting pixels at the beginning of training and gradually transitions to predicting latent features at deeper layers as we progressively freeze the network. 

\vspace{1mm}
\noindent \textbf{Dataset.} We use a dataset of 174M web videos, each 900-frames long on average. We sample 16 frame clips with a stride of 2 and randomly crop to $224\times 224$ after resizing the smallest side to $1.15\times$ the resolution. We additionally apply random cropping and horizontal flipping. We use a patch size of $2 \times 16 \times 16$ and mask $95\%$ of the patches. As mentioned above in \cref{sec:methodology}, we experiment with dropping some of the input patches when computing targets for latent loss. In one set of experiments, we compute \textit{latent loss} on a random 5\% subset of all patches, while in another set of experiments, we do not drop any patches and use all of them when computing the latent loss. Note that, in contrast, for \textit{pixel loss}, we always use all of the patches. 

\vspace{1mm}
\noindent \textbf{Training details.} We train 4DS-\Model on 1 billion video clips on 256 TPUs-v6 using the AdamW optimizer \citep{loshchilov2019adamw}. We use a cosine learning rate schedule with a warmup and decay. We find it beneficial to have mini-warmups whenever we switch from one target to the next (see \cref{sec:results} for more details.) For efficiency, we train the 4DS models in \textit{bfloat16} precision except for the layer normalization, softmax and loss computations. All weights are stored in \textit{float32} and we use data and model parallelism. Additional details are provided in the supplementary material.

We follow the 4DS decoder architecture, using the last 4 layers of the Vision Transformer backbone. For progressive freezing, we determine the schedule following this simple heuristic: start the freezing when the norms of the first few layers seems to plateau (or start getting smaller when using weight decay). Then freeze layers one by one until we freeze 3/4 of all the layers at the end of training. For example, for the 4DS ViT-G model, this results in a schedule where we train with pixel loss for 160K steps and freeze 1 layer every 10K steps until we freeze 32 layers at the end of training. For rotary positional embeddings, we use a max wavelength of 10,000 and use 10\%, 25\%, and 25\% of the input vector for encoding time, height, and width respectively.

\subsection{Extending \Model to V-JEPA}
\label{sec:vjepa_method}

To demonstrate the generality of our approach beyond the MAE methodology, we also show that \Model can be applied to video models that already leverage latent prediction. Specifically, we choose the highly popular V-JEPA~\cite{bardes2023vjepa} model. This model is also trained on masked videos but predicts latent features encoded by a teacher network rather than pixels. The teacher network here uses an exponentially weigthed moving average (EMA) of the student network to provide stable targets for prediction and is crucial for avoiding representation collapse as shown by earlier work \cite{grill2020byol}.

We apply \Model on V-JEPA by predicting the first layer activations from the teacher network at the beginning of training and again gradually transitioning to predicting deeper layer activations as we freeze more and more layers of the encoder. For model architecture and training setup, we follow the original V-JEPA implementation as much as possible (see the supplementary material for more details) and use a ViT-L backbone for the encoder while using a separate $12$ layer transformer for the decoder. For training our baseline V-JEPA and \Model versions, we use the Kinetics700 dataset \cite{kay2017kineticshumanactionvideo} and train on 560 million video clips following the original V-JEPA paper. Note our training dataset is different from the one used in the original paper, which combined multiple Kinetics datasets into a single larger dataset. All of the data preprocessing setup is the same with the 4DS models, except in addition to the augmentations described for the 4DS models, we also use  color jittering for training the V-JEPA models.
We train on 256 TPUs-v3 using the AdamW optimizer \citep{loshchilov2019adamw} with a cosine learning rate schedule with a warmup and decay. We determine the freezing schedule following the simple heuristic described above, which results in a schedule where we train the full unfrozen model on predicting first layer features for 100K steps and freeze 1 layer every 6K steps afterward until we freeze all 24 layers at the end of training.

\subsection{Evaluation}

Since we are interested in the quality of the representations learned by \Model, we do not use fine-tuning for evaluations, but rather freeze the model once it is trained, and train a task-specific readout on a number of examples for each task. Given a video clip for evaluation, we add positional embeddings and feed it to our model without any masking to get the patch embeddings that are used for readouts. We get the patch embeddings from the layer positioned at 95\% of the total network depth (e.g., for a 48-layer network, this corresponds to layer 45.). We use attention-based readouts, which in our experiments, performed much better than linear readouts. Depending on the task, we use a number of learned latent tokens to cross-attend to the patch embeddings and finally project to the output space of the evaluation task (see the supplementary material for more details). We train the readouts using 1.28M examples and AdamW optimizer on the evaluation task.

\vspace{1mm}
\noindent \textbf{Tasks.} We evaluate on a set of both low and high-level visual tasks from the 4DS perception tasks suite \cite{carreira2024scaling}: (1) action classification on Kinetics700~\citep{kay2017kineticshumanactionvideo} (2) action classification on SSv2~\cite{goyal2017something}, both of which require  higher-level, spatio-temporal semantic understanding (3) monocular depth estimation in ScanNet~\cite{dai2017scannet}, which test lower-level 3D perception. We provide full detail about all the tasks in the supplementary material.

\subsection{Baselines}
As mentioned in~\cref{sec:training}, we apply \Model on strong representatives of pixel and latent prediction approaches for video representation learning. Hence we compare \Model to 4DS ~\cite{carreira2024scaling} model, and V-JEPA~\cite{bardes2023vjepa} model, as representatives of the pixel and latent prediction approaches respectively. All models use a ViT backbone and tokenize videos into patches of size $2 \times 16 \times 16$.

%% file: sec/4_results.tex
\section{Results}
\label{sec:results}

\subsection{Progressive freezing for more efficient MAE training}

Leveraging the convergence observations described in Sec. \ref{sec:layer_convergence} and shown in Fig. \ref{fig:convergence_step}, we now show that we can make MAE training more efficient without losing performance simply with progressive layer-freezing. For this experiment, we use a ViT-G backbone trained on 250M examples (same as for the ablations in Sec. \ref{sec:ablations}). 

\begin{figure}[htbp]
    \centering
    \vspace*{12pt}
    \includegraphics[width=\linewidth]{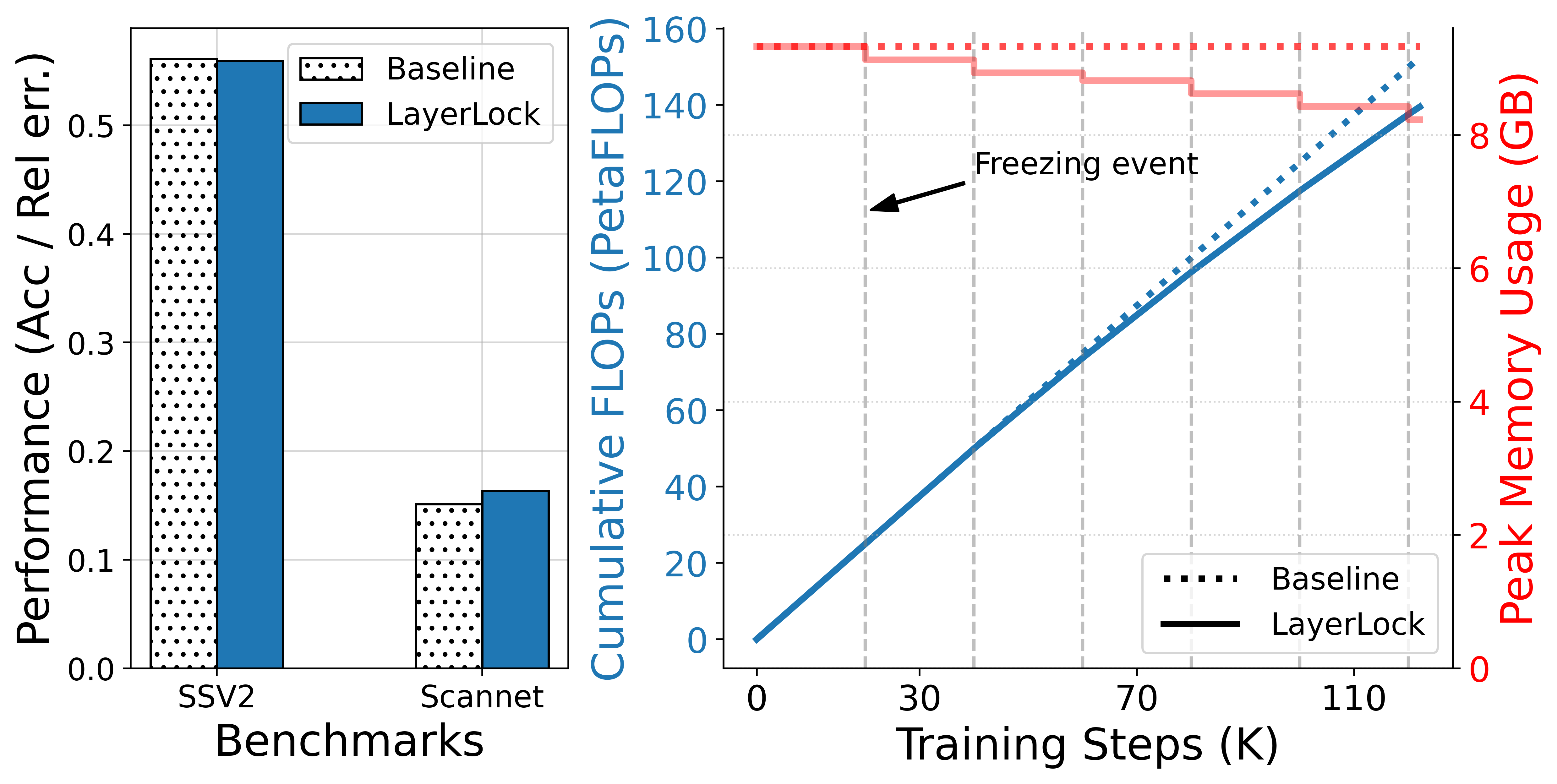}
    \caption{\textbf{Progressive freezing of layers saves on total training cost and peak memory utilization without loss in performance.} \textbf{(Right)} Cumulative training cost (PetaFLOPs) and peak memory usage at each step (GiB) are shown for the baseline (unfrozen) and our method. Freezing events are indicated by gray vertical lines. \textbf{(Left)} Performance of baseline unfrozen MAE (dotted) vs. progressive freezing MAE on SSV2 (accuracy) and ScanNet depth pred (rel. error).}
    \label{fig:flops_mem}
\end{figure}

Fig. \ref{fig:flops_mem} (Left) highlights that we can progressively freeze (filled bars) with negligible loss in performance compared with the baseline (dotted bars) on two tasks (56.1\% vs. 56.0 accuracy on SSv2, and 0.15 vs. 0.16 relative error on ScanNet). Moreover, this competitive performance is achieved with 9\% fewer total FLOPs and 16\% less peak memory usage as seen in Fig. \ref{fig:flops_mem} (Right). We note that while this experiment was done at a shorter schedule, these efficiency gains only get larger with longer schedules- at 1B training examples, FLOP efficiency gains go up to 19\%. 

\subsection{\Model vs. Baselines}

\begin{table*}[t]
\centering
\centering
\vspace{1mm}
\caption{\Model improves training of both pixel and latent prediction based video models when evaluated on action recognition and depth prediction tasks. \textbf{Bold} indicates best performance within the group. All results are obtained using the same evaluation protocol (see text for details). ScanNet values are multiplied by 10 for better readability.}
\renewcommand{\arraystretch}{0.9}
\begin{tabular}{lcc|cc|c}
\toprule
\textbf{Model} & \textbf{LayerLock} & Size (M) & \textbf{SSv2} & \textbf{K700-2020} & \textbf{ScanNet}   \\ 
& & & ($\uparrow$) &($\uparrow$)&($\downarrow$) \\
\toprule
4DS-G (MAE) & \xmark & 1,848  & 63.1  &   52.1 &  1.02  \\
\rowcolor{DnCBG} 4DS-G (MAE) & \checkmark  & 1,868  & \textbf{66.1}  &  \textbf{56.3}  & \textbf{1.00}  \\
\midrule
4DS-e (MAE) & \xmark & 3,811  & 64.6  &  54.4 &  \textbf{0.94} \\
\rowcolor{DnCBG} 4DS-e (MAE) & \checkmark   & 3,818  & \textbf{67.1}  &  \textbf{57.9}  & 0.98  \\
\midrule
V-JEPA-L  & \xmark    & 235 & 52.1 & 42.5 & 1.57 \\ 
\rowcolor{DnCBG} V-JEPA-L  & \checkmark  & 303 & \textbf{57.0} & \textbf{43.5} & \textbf{1.51} \\ 
\bottomrule
\end{tabular}
\label{tab:model_comparison}
\vspace{-0.3cm}
\end{table*}

We begin by looking at how well our proposed model \Model performs compared to baseline video models in the literature. Our aim here is to show that applying our \Model recipe to both pixel and latent prediction approaches leads to better video representations. We evaluate each model under the same evaluation protocol described in \cref{sec:methodology} and report results in \cref{tab:model_comparison}. 

We use the following metrics for each evaluation task. For SSv2 and Kinetics700 action classification, we report Top-1 accuracy as a percentage. For ScanNet depth prediction, we use Absolute Relative Error (Lower is better and the minimum value is 0).

In~\cref{tab:model_comparison}, we show how \Model compares with the baseline 4DS MAE approach at two different model scales (G and e). Compared to the baseline models, those that use our \Model approach improve significantly on both action recognition tasks while maintaining or improving slightly on depth prediction. 

To further demonstrate the generality of our method, we also show that \Model can also be applied to the latent prediction method, V-JEPA. We see that even in this case, \Model leads to significant improvements both on SSv2 and K700-2020 action recognition tasks, and ScanNet depth estimation task. In sum, these results demonstrate that \Model can be a general recipe for improving the video representations learned by both pixel and latent prediction approaches.

\subsection{Ablations}
\label{sec:ablations}
We run an extensive series of ablation studies to understand why \Model performs better than a vanilla MAE model. Unless otherwise specified, for ablation experiments, we run a  ViT-G model on 250M examples and evaluate on SSv2 action classification and ScanNet depth prediction.

\vspace{1mm}
\noindent \textbf{Efficient latent prediction via patch subselection.}
The results in \cref{tab:model_comparison} show the effectiveness of \Model, but as described earlier, we also explore increasing the efficiency of the method by restricting the number of patches on which the latent loss is calculated. We see in \cref{tab:ablation-patch-percent} that even with \textit{only 5\% of patches} used for latent loss, \Model leads to performance improvements in action classification tasks while only showing a modest decrease in depth estimation performance. It will be interesting to explore this performance vs. compute trade-off in a more granular way in future work.

\begin{table}
\centering
\caption{Ablation: We can increase the efficiency of \Model by using just 5\% of patches for the latent-loss. This under-performs our best model but still beats the baseline significantly on SSv2. For this ablation only we use the same training settings as in 
\cref{tab:model_comparison}}.

\begin{tabular}{lc|c|c}
\toprule
\textbf{Model} & \textbf{\makecell{Latent Loss \\ \% Patches}} & \textbf{SSv2} & \textbf{ScanNet} \\ 
& & ($\uparrow$) &($\downarrow$) \\
\midrule
4DS-G (MAE) & - & 63.1 & 1.02 \\
4DS-G LayerLock & 5\%   & 64.9  & 1.12 \\
4DS-G LayerLock & 100\% & 66.1  & 1.00  \\
\bottomrule
\end{tabular}
\label{tab:ablation-patch-percent}
\end{table}

\vspace{1mm}
\noindent \textbf{MAE with latent losses collapses.} We've demonstrated that \Model benefits from two components: progressive freezing and switching from pixel reconstruction to latent prediction losses, but one might ask, can we just add latent losses to the traditional MAE? In \cref{tab:ablation_latent}, we test this using the 4DS-H MAE model (600M parameters) on 100M examples (due to resource constraints). We train on a weighted sum of latent losses (from intermediate layers) and the pixel loss. We experiment with both constant weight for the latent losses and a cosine schedule over latent loss weight. In all cases, we see serious signs of collapsing of the learned representations. This demonstrates the necessity of progressive freezing to avoid such collapse when introducing latent prediction losses. 

\begin{table}
\centering
\caption{Ablation: Adding latent losses to the MAE paradigm without freezing leads to representation collapse.}
\begin{tabular}{l|c|c}
\toprule
\textbf{Model} & \textbf{SSv2} & \textbf{ScanNet} \\ 
& ($\uparrow$) &($\downarrow$) \\
\midrule
4DS-H  & 50.1 & 0.19 \\
4DS-H + latent (const) &  3.7 & 0.38 \\
4DS-H + latent (cosine) &  5.6 & 0.37 \\
\bottomrule
\end{tabular}
\label{tab:ablation_latent}
\end{table}

\vspace{1mm}
\noindent \textbf{3D rotary positional embeddings.} As mentioned in \cref{sec:methodology}, we use a novel 3D version of rotary positional embeddings (RoPE). Our experiments (shown in \cref{tab:ablation_rope}) show that our RoPE variant consistently improves performance for both the baseline and our \Model model, yielding, for instance, ~2.5\% improvement in SSv2 classification performance.

\begin{table}
\centering
\caption{Ablation: Adding 3D RoPE embeddings improves baseline models independent of \Model. In concert with \Model, we get further improvements.}
\begin{tabular}{lc|c|c}
\toprule
\textbf{Model} & \textbf{3D-RoPE} & \textbf{SSv2} & \textbf{ScanNet} \\ 
& & ($\uparrow$) &($\downarrow$) \\
\midrule
4DS-G   & \xmark & 56.1 & 0.15 \\
4DS-G & \checkmark  & 58.9 & 0.13 \\
4DS-G \Model & \checkmark & 60.1 & 0.13 \\
\bottomrule
\end{tabular}
\label{tab:ablation_rope}
\end{table}

\vspace{1mm}
\noindent \textbf{Single vs. multiple targets.} At any point in training, \Model predicts only a single target (either pixels or an intermediate layer). An alternative approach would be to keep the previous prediction targets every time we switch to a new target and simply sum the losses from each target as we add new training targets throughout training. This might be beneficial because the model is always trained on pixel loss in this setting and has less risk of diverging and forgetting how to predict pixels, as opposed to the single target approach. Notably, our experiments (reported in~\cref{tab:ablation_multiple}) reveal that the simpler approach we use in \Model, of always keeping the latest target, and training on a \textit{single loss} (instead of pixel and all latent losses together) results in an equally well performing model.

\begin{table}
\centering
\caption{Ablation: Single prediction targets from the latest frozen layer are sufficient in the \Model setup when compared with more complex multi-target prediction.}
\begin{tabular}{l|c|c}
\toprule
\textbf{LayerLock Target Choice} & \textbf{SSv2} & \textbf{ScanNet} \\ 
& ($\uparrow$) &($\downarrow$) \\
\midrule
Single (latest) target & 55.5 & 0.16 \\
Multiple targets & 55.4 & 0.16 \\
\bottomrule
\end{tabular}
\label{tab:ablation_multiple}
\end{table}

\vspace{1mm}
\noindent \textbf{Latent loss warmup.} As mentioned in \cref{sec:methodology}, every time we switch to predicting a new intermediate layer, we do a mini learning rate warmup where we gradually increase the learning rate. Our experiments (shown in \cref{tab:ablation_warmup}) indicate that this leads to a small but consistent improvement in performance (e.g., ~1\% on SSv2 action classification).

\begin{table}
\centering
\caption{Ablation: Effect of latent loss warmup.}
\begin{tabular}{l|c|c}
\toprule
\textbf{Latent Loss Warmup Steps} & \textbf{SSv2} & \textbf{ScanNet} \\ 
& ($\uparrow$) &($\downarrow$) \\
\midrule
No warmup & 58.7 & 0.14 \\
1K steps & 59.8 & 0.15 \\
3K steps & 59.4 & 0.15 \\
\bottomrule
\end{tabular}
\label{tab:ablation_warmup}
\end{table}

\vspace{1mm}
\noindent \textbf{Freezing schedule.} In our experiments, we use a freezing schedule defined by the following parameters: \textit{freezing start}: number of initial steps before freezing, \textit{freezing interval}: number of steps between each freezing event, \textit{layer jump}: number of layers to freeze at each freezing event, and \textit{target layers}: the layers to compute latent losses for. In this study, we vary each of these parameters to understand how they affect downstream performance. Due to resource constraints, we run these experiments on a ViT-B model trained on 50M examples (see \cref{tab:ablation_schedule}). First, as we vary the freezing start step, we see that fewer steps tend to lead to worse performance, suggesting that in these cases the model is not trained on pixel loss long enough (i.e., the early layers have not converged yet). Secondly, we look at how the number of layers we freeze at each freezing event (i.e, layer jump) affects performance. As we freeze more layers, we see significant drops in performance. This is likely due to more layers being frozen for longer as we increase the layer jump. Finally, we look at how the freezing interval affects performance. Note that as we increase the freezing interval, we need to adjust the layer jump to make sure we freeze the same proportion of layers in the model in each case. We see that increasing the freezing interval leads to worse performance, with the shortest freezing interval (2K steps) giving the best results. This suggests that gradual freezing of layers is more effective, even though in this case earlier layers are kept frozen for longer.

\begin{table}
\centering
\caption{Ablation: Effect of freezing schedule. Baseline model uses a schedule with start=6K, interval=4K, jump=2, targets=(1,3,5,7).}
\begin{tabular}{l|c|c}
\toprule
\textbf{Model} & \textbf{SSv2} & \textbf{ScanNet} \\ 
& ($\uparrow$) &($\downarrow$) \\
\midrule
4DS-G \Model & 38.3 & 0.24 \\
\addlinespace[0.3em]
\hspace{3mm}   start=2K & 36.0 & 0.25 \\
\hspace{3mm}   start=4K & 37.1 & 0.25 \\
\addlinespace[0.3em]
\hspace{3mm}   jump=3, targets=(2, 5, 8, 11) & 36.0  & 0.25 \\
\hspace{3mm}   jump=4, targets=(3, 7, 11) & 33.1 & 0.26 \\
\addlinespace[0.3em]
\hspace{3mm}   interval=2K, jump=1 & 40.0 & 0.24 \\
\hspace{3mm}   interval=8K, jump=4 & 36.2 & 0.25 \\
\hspace{3mm}   interval=16K, jump=8 & 28.5 & 0.28 \\
\bottomrule
\end{tabular}
\label{tab:ablation_schedule}
\end{table}

%% file: sec/5_conclusion.tex
\section{Conclusions}
\label{sec:conclusions}

In this paper, we first presented an observation: when training  masked-autoencoding models with ViT architectures, earlier layers in the network converge earlier than later ones. This motivated our approach \Model, that progressively freezes layers according to a freezing schedule and saves compute and memory while reaching the same final performance.

Secondly, following this observation, we have presented a new self-supervised video representation learning approach that trains on a dynamically evolving prediction target, starting from low-level shallow features and gradually transitioning to increasingly higher-level deeper features throughout training, as we progressively freeze the model's layers. This results in an effective representation learning technique that is more efficient than pixel prediction approaches like MAE and yet is more stable than latent prediction approaches, avoiding their common issue of representation collapse. As shown in our main results and extensive ablations, \Model leads to improvements in high-level visual downstream tasks like action classification, while largely maintaining performance on low-level tasks like depth estimation.

Looking ahead, we are excited about the potential compute and memory saved by progressive freezing, as it opens new avenues for scaling to longer videos, larger resolutions and even deeper models.

%% file: sec/6_supplementary.tex
\clearpage
\setcounter{page}{1}
\maketitlesupplementary

\appendix

In this supplementary material we provide additional details on pretraining, evaluation protocol, tasks, readouts and baselines. We also present details of the layer convergence analysis that partly motivates our approach.

\section{Additional details}
\label{sec:supp_details}

\subsection{Layer Convergence Analysis}
We describe here the setup used for the convergence analysis shown in Fig. \ref{fig:convergence_step}. We start with the basic setup used by the 4DS MAE models in \cite{carreira2024scaling}. However, due to the large number of experiments required, we use a small model with a relatively short training schedule. 

Specifically, we start with a modified ViT-S encoder that has 16 layers. We train the baseline and all layer-freezing ablations for 14K training steps, with batch size 2048. We use the same optimizer settings described below but with a learning rate of 1e-4. 

To remove noise in the loss convergence calculations, we take the ``final loss'' as the average loss in the final 1000 steps of training. 

\subsection{Pretraining}

In \cref{tab:pretraining_hparams_4ds} we provide the hyperparameters used to train our 4DS models. As mentioned in the main text, these models were trained on 1B examples. Note during optimization, we do not apply weight decay on layer norm parameters and biases of linear layers.

\cref{tab:pretraining_hparams_vjepa} provides the hyperparameters used to train our V-JEPA models. These largely follow the hyperparameters in the original paper (scaled for the slightly smaller batch size of 2048 as opposed to 3072) with all models trained on 560M examples from the Kinetics700 dataset \cite{kay2017kineticshumanactionvideo}. Note this is different from the training dataset in the original paper, which combined multiple Kinetics datasets into a single larger dataset. Also note that even though our training setup is the same as the original paper, we use the same evaluation setup for all models (which is different from the original paper) and hence our V-JEPA numbers are not directly comparable with the ones in the original paper. In contrast to the 4DS models, we use learned positional embeddings instead of RoPE for V-JEPA models, and we do all computations in \textit{float32} precision. 

Note for V-JEPA, we still use an EMA of the encoder to compute the targets. This is not strictly necessary to avoid representation collapse as the progressive freezing employed by \Model helps avoid such a problem. However, our initial experiments have shown that using an EMA target network results in better downstream performance. 

\begin{table}[h]
\centering
\begin{tabular}{l c}
\hline
\textbf{Hyperparameter} & \\
\hline
Num. training steps & 488,282 \\
Input resolution & 224$\times$224 \\
Learning rate & 3e-4 \\
Warmup steps & 10,000 \\
\#N decoding layers & 4 \\
Patch size & 2$\times$16$\times$16 \\
Minimum resize factor & 1.15 \\
Batch size & 2048 \\
MAE masking ratio & 0.95 \\
AdamW weight decay & 0.05 \\
AdamW b1 & 0.90 \\
AdamW b2 & 0.95 \\
\midrule
Num. of steps before first freezing & 160,000 - 200,000 \\
Num. steps between freezing & 10,000 \\
Num. layers to freeze & 1 \\
Target layers & (1, 2, 3, \dots, 32) \\
\midrule
RoPE max. wavelength & 10,000 \\
RoPE T, H, W proportions & 10\%, 25\%, 25\% \\
\hline
\end{tabular}
\caption{Pretraining hyperparameters for 4DS ViT-G and ViT-e models trained on 1B examples. Note all hyperparameters are the same for both models except the number of steps before freezing. Minimum resize factor controls how much a video's minimum side gets resized before cropping as function of input resolution.}
\label{tab:pretraining_hparams_4ds}
\end{table}

\begin{table}[h]
\centering
\begin{tabular}{l c}
\hline
\textbf{Hyperparameter} & \\
\hline
Num. training steps & 262,501 \\
Input resolution & 224$\times$224 \\
Initial learning rate & 1.3e-4 \\
Learning rate & 4.17e-4 \\
End learning rate & 6.6e-7 \\
Warmup steps & 90,000 \\
Patch size & 2$\times$16$\times$16 \\
Stride & 4 \\
Minimum resize factor & 1.15 \\
Batch size & 2048 \\
Target network EMA coef. & 0.998 \\
AdamW weight decay & 0.04 $\to$ 0.4 \\
\midrule
Multiblock masking parameters & \\
\hspace{0.2em} Num. blocks & 8 \\
\hspace{0.2em} Block area range & (0.3, 0.3) \\
\hspace{0.2em} Aspect ratio range & (0.75, 1.50) \\
\midrule
Num. of steps before first freezing & 100,000 \\
Num. steps between freezing & 6,000 \\
Num. layers to freeze & 1 \\
Target layers & (1, 2, \dots, 24) \\
\hline
\end{tabular}
\caption{Pretraining hyperparameters for V-JEPA trained on 560M examples.}
\label{tab:pretraining_hparams_vjepa}
\end{table}

\subsection{Ablation studies}

Due to resource constraints,  we used models of different sizes trained on different number of examples for ablations. In \cref{tab:ablation_hparams}, we provide the hyperparameters used to train these.

\begin{table*}[h]
\centering
\begin{tabular}{l c c c c}
\hline
\textbf{Hyperparameter} & \textbf{ViT-G, 250M} & \textbf{ViT-G, 56M} & \textbf{ViT-H, 100M} & \textbf{ViT-B, 50M} \\
\hline
Learning rate & 1e-4 & 1e-4 & 1e-3 & 3e-4 \\
Warmup steps & 5,000 & 5,000 & 1,000 & 2,000 \\
Batch size & 2048 & 128 & 8096 & 512\\
Weight decay & No & No & No & 1e-3\\
\midrule
Num. of steps before first freezing & 25,000 & 19,000 & -- & 6,000 \\
Num. steps between freezing & 20,000 & 30,000 & -- & 4,000 \\
Num. layers to freeze & 5 & 5 & -- & 2\\
Target layers & (4, 8, 12, 16) & (4, 8, 12, 16, $\dots$, 44) & -- & (1, 3, 5, 7) \\
\hline
\end{tabular}
\caption{Hyperparameters for models used in ablation studies. Only hyperparameters that are different than \cref{tab:pretraining_hparams_4ds} shown.}
\label{tab:ablation_hparams}
\end{table*}

\subsection{Evaluation}

Our evaluation closely follows the setup in \cite{carreira2024scaling}, which we discuss in summary here. See the supplementary material in \cite{carreira2024scaling} for more details.

As mentioned in the main text, we evaluate all models by training a readout module on top of frozen features. We use 1.28M training examples for each task with a batch sie of 32. We sweep over learning rates (1e-4, 3e-4, 1e-3) and readout depth fraction (0.25, 0.5, 0.75, 0.85, 0.95, 1.0) and report the best results. We use a cosine schedule for the learning rate with linear warmup of 1K steps and a decay to 1e-7. We use a similar cross-attention readout architecture for all tasks, where a set of learned tokens cross-attend to frozen features. A summary of the readout configurations and number of parameters in each case is provided in~\cref{table:readout_modules}.

\begin{table*}[t]
\centering
\begin{tabular}{c|l c}
\hline
\textbf{Eval} & \textbf{Architecture} & \textbf{Number of Params} \\ \hline
SSv2 Classification & 
\begin{tabular}[c]{@{}l@{}}
\texttt{LearnedQueries(num\_channels=768)} \\
\\
\texttt{CrossAttention(qkv\_size=768, num\_heads=12)} \\
\\ 
\texttt{Linear(output\_size=174)}
\end{tabular} & 7M \\ \hline
Kinetics Classification & 
\begin{tabular}[c]{@{}l@{}}
\texttt{LearnedQueries(num\_channels=1024)} \\
\\
\texttt{CrossAttention(qkv\_size=1024, num\_heads=16)} \\
\\ 
\texttt{Linear(output\_size=700)}
\end{tabular} & 7M \\ \hline
ScanNet Depth Prediction & 
\begin{tabular}[c]{@{}l@{}}
\texttt{LearnedQueries(num\_channels=1024)} \\
\\
\texttt{CrossAttention(qkv\_size=1024, num\_heads=16)} \\
\\
\texttt{Linear(output\_size=128)}
\end{tabular} & 18M \\ \hline
\end{tabular}
\caption{Configurations and number of parameters of cross-attention-based readout modules used in this paper, for different tasks. Note that the number of parameters are given for the case of a ViT-L backbone, that has 1024-channel outputs.}
\label{table:readout_modules}
\end{table*}

In the following, we provide a short description of each evaluation task.

\subsubsection{Something-Something v2 action classification}
The SSv2 action classification dataset contains 220,000 videos with duration ranging from 2 to 6 seconds at 12fps. Videos contain 174 human actions with everyday objects.

\vspace{2mm}
\noindent \textbf{Task definition.} Given a video clip of 16 frames of resolution 224x224 with stride 2, the model is tasked to predict an action class. Top-1 accuracy is used to measure the performance.

\begin{table}[t]
\centering
\begin{tabular}{l|l}
\hline
 Brightness & delta in [-0.125, 0.125] \\ \hline
 Saturation & factor in [0.6, 1.4] \\  \hline
 Contrast & factor in [0.6, 1.4] \\  \hline
 Hue & delta in [-0.2, 0.2]  \\ \hline
\end{tabular}
\caption{Hyper-parameters for color augmentation used when training readout heads on the SSv2 task. Deltas are added to the corresponding channel, while factors multiply the corresponding channel.}
\label{tab:colour_augment}
\end{table}

\vspace{2mm}
\noindent \textbf{Readout details.} The cross-attention readout module uses 768 parameters with 12 heads and a single learned query to predict logits for 174 classes. In training, we resize the shorter size of the video to 239 and take random temporal crop of shape 224x224 from it. We use colour augmentation with 0.8 probability of randomly adjusting the brightness, saturation, contrast and hue (see \cref{tab:colour_augment}), and a 0.1 probability of converting to grayscale. In test time we take one 224x224 central crop from the video without any colour augmentation.

\subsubsection{Kinetics700 action classification}
The SSv2 action classification dataset contains 545,317 10s video clips from 700 action classes.

\vspace{2mm}
\noindent \textbf{Task definition.} Given a video clip of 16 frames of resolution 224x224 with stride 2, the model is tasked to predict an action class. At test time 7 equally spaced clips are passed through the trained classifier and their softmax scores are averaged to get predictions. Top-1 accuracy is used to measure the performance.

\vspace{2mm}
\noindent \textbf{Readout details.} The cross-attention readout module uses 1024 parameters with 16 heads and a single learned query to predict logits for 700 classes.

\subsubsection{ScanNet depth estimation}

ScanNet~\cite{dai2017scannet} is a video dataset captured in various indoor environments, containing rich annotations for 3D camera poses, surface reconstructions, and instance-level semantic segmentations. Data was obtained through an RGB-D capture system that produces depth. RGB frames have 1296x968 resolution while depth frames have 640x480 resolution. There are 1201 videos in the train split, 312 videos in the validation split, and 100 videos in the test split. We use the train and validation splits of ScanNet for this paper.

\noindent \textbf{Task definition.} We feed the models 16 RGB frames while adding readout heads on top to output depth for each input frame. We scale the images to the (0, 1) range and mask out target depth values outside of (0.001, 10) meters. We perform random cropping and left-right flipping during training and take a center crop during testing. 

\noindent \textbf{Evaluation metrics.} We follow prior work on monocular depth estimation and report the mean of the absolute relative error (AbsRel) \citep{steenkiste2024moving,yang2024depth,ranftl2021vision} which is computed as $|d^{*}-d|/(d+\epsilon)$ where $d^{*}$ is the predicted depth values, $d$ is the ground truth depth. 

\noindent \textbf{Readout details.} We use a cross-attention readout head with 1024 parameters and 16 heads with one learned query for each spatio-temporal patch in the input video. We use a patch size of $2 \times 8 \times 8$ and predict 128 ($=2*8*8$) depth values for each patch, one for each pixel. We use an L2 loss.